\pdfoutput=1
\documentclass[11pt]{article}

\usepackage[preprint]{acl}

\usepackage{times}
\usepackage{latexsym}
\usepackage{graphicx}
\usepackage{listings}

\usepackage[T1]{fontenc}
\usepackage[utf8]{inputenc}

\usepackage{microtype}

\usepackage{inconsolata}

\title{GPT is Not an Annotator: The Necessity of Human Annotation in Fairness Benchmark Construction}

\author{
  Virginia K. Felkner \\
  Information Sciences Institute\\
  University of Southern California\\
  \texttt{felkner@isi.edu} \\
  \And
  Jennifer A. Thompson \\
  Jewish Studies Program \\
  California State University, Northridge\\
  \texttt{jennifer.a.thompson@csun.edu} \\
  \AND
  Jonathan May \\
  Information Sciences Institute\\
  University of Southern California\\
  \texttt{jonmay@isi.edu} \\
  }

\begin{document}
\maketitle
\begin{abstract}
Social biases in LLMs are usually measured via bias benchmark datasets. Current benchmarks have limitations in scope, grounding, quality, and human effort required. Previous work has shown success with a community-sourced, rather than crowd-sourced, approach to benchmark development. However, this work still required considerable effort from annotators with relevant lived experience. This paper explores whether an LLM (specifically, GPT-3.5-Turbo) can assist with the task of developing a bias benchmark dataset from responses to an open-ended community survey. We also extend the previous work to a new community and set of biases: the Jewish community and antisemitism. Our analysis shows that GPT-3.5-Turbo has poor performance on this annotation task and produces unacceptable quality issues in its output. Thus, we conclude that GPT-3.5-Turbo is not an appropriate substitute for human annotation in sensitive tasks related to social biases, and that its use actually negates many of the benefits of community-sourcing bias benchmarks.
 
\end{abstract}

\section{Introduction}
Though seemingly ubiquitous, large language models (LLMs) still treat users unequally and exhibit harmful social biases \cite{weidinger-taxonomy, shelby-taxonomy}.\footnote{Clear and explicit definitions of the terms \textit{bias} and \textit{harm} are essential for productive discussion of AI fairness \cite{blodgett-etal-2020-language}. 
For purposes of this paper, we define \textit{bias} as ``substantially differing treatment of a marginalized group relative to a dominant group that replicates existing social stereotypes about the marginalized group'' and \textit{harm} as ``physical, psychological, financial, or professional events that affect a person in perceived negative way.'' } 
Quantitative LLM bias measurement is a necessary first step to understanding and mitigating bias-related harms of AI systems. 
Measurement is essential because it allows model creators to understand potential fairness issues with their models, downstream users to compare models and choose those that are relatively fair in their use context, and fairness researchers to determine whether debiasing methods are effective.  

The current standard for bias measurement in LLMs is paired sentence bias benchmarks, which consist of pairs of similar sentences and generally rely on comparing the model's probability of predicting the stereotypical sentence to the probability of predicting a contrasting sentence. 
There are significant quality and grounding issues with most current benchmarks, especially those developed via crowd-sourcing. 
Current methods for community-sourced benchmark development, which mitigate some of the problems with crowd-sourcing, require significant human effort for annotation of survey responses. 
This work is time-consuming; in an unfunded, community-led benchmark development effort, this is either cost-prohibitive or requires asking annotators to work for free. This annotation also places a significant psychological burden on annotators. 

In order to maintain the usefulness and participatory nature of community-sourced bias benchmarks while reducing the financial and psychological costs, we tested \textbf{model-assisted harm extraction.} The main contributions of our work are as follows:\footnote{Our code and data are available at \url{https://github.com/katyfelkner/winosemitism}}
\begin{itemize}
    \item We introduce WinoSemitism, a community-sourced  benchmark for antisemitism, generalizing method of \citet{felkner-etal-2023-winoqueer}. 
    \item We create GPT-WinoQueer and GPT-WinoSemitism, which are versions of the WQ and WS datasets annotated by GPT-3.5-Turbo instead of human experts.
    \item We provide a thorough quantitative and qualitative comparison of human-annotated and model-annotated datasets, finding significant quality issues with model annotations. 
\end{itemize}

\section{Related Work}

The current standard for bias measurement in LLMs is paired sentence bias benchmarks, such as StereoSet \cite{nadeem-etal-2021-stereoset} and CrowS-Pairs \cite{nangia-etal-2020-crows}.
These benchmark datasets consist of pairs of stereotypical (e.g. e.g ``Women are bad at math.'') and counter-stereotypical (e.g. ``Men are bad at math.'') sentences. 
Metrics vary slightly between benchmarks, but generally rely on comparing the model's probability of predicting the stereotypical sentence to the probability of predicting the other sentence. 

While paired sentence benchmarks are important tools for understanding and mitigating bias, the current benchmarks have several important weaknesses.
First, most benchmarks attempt to be general-purpose, covering a variety of bias axes (e.g. race, gender, nationality,  religion, etc.) with the intention that downstream users need only test their models on one benchmark. 
However, even the most extensive benchmarks leave out many biases that exist and are harmful in the real world. 
Even when they cover several bias axes, benchmarks are often oversimplified and lacking nuance along each axis. For example, \citet{nangia-etal-2020-crows} include homophobia in their benchmark, but treat LGBTQ+ identity as a binary attribute rather than a complex set of related communities. Similarly, \citet{nadeem-etal-2021-stereoset} include religion as one of their bias axes, but only consider Christianity, Islam, and Hinduism, ignoring antisemitism and other religious biases. LLM benchmarking work on antisemitism is limited; however, there is closely related work on antisemitism in the context of hate speech detection \cite{subverting2021, Jikeli2021DetectingAM}.

Second, many existing benchmarks have serious quality control issues, including ungrammatical or nonsensical sentences, mismatching of stereotypes to target groups, and vague or nonspecific stereotypes \cite{blodgett-etal-2021-stereotyping}.
Finally, most existing benchmarks are insufficiently \textit{grounded,} meaning that the stereotypes and biases described in the benchmark may not exist in reality and are not well-aligned with the affected communities' opinions on what stereotypes are harmful and what constitutes unacceptable or undesirable LLM behavior \cite{blodgett-etal-2020-language}.

Grounding issues often stem from the method of collecting stereotypes to include in bias benchmarks.
Many benchmark creators use \textit{crowd-sourced} stereotypes, usually from crowdworkers on Amazon Mechanical Turk, who have varying degrees of knowledge and personal experience with social biases.
In contrast, recent work \cite{felkner-etal-2023-winoqueer} had success with \textit{community-sourced} bias benchmarks in which stereotypes were derived from a large-scale online survey of members of the affected community. 
This yielded a large, specific, well-grounded, and high-quality bias benchmark dataset for harms affecting a specific community. 

Community-sourcing, while a significant improvement over crowd-sourcing in many respects, has its own problems. 
Current methods require survey responses to be annotated by hand, ideally by researchers who are themselves members of the communities surveyed. 
This annotation relies heavily on the lived experience and subjective opinions of annotators to extract \textit{attested harm predicates} from survey responses, which are then inserted into template sentences to create the stereotypical and counter-stereotypical sentences in the benchmark.
Human annotation by researchers with relevant lived experience is time-consuming, and therefore expensive and often cost-prohibitive for grassroots, community-based efforts. 

Additionally, the annotation is psychologically taxing on annotators. Annotators spend hours reading detailed descriptions of violence and hatred toward their own communities. This is often triggering for annotators who have first-hand experiences with such harm, and it is exhausting and depressing even for those who have somewhat less painful experience. The situation is somewhat analogous to content moderation, in which repeated exposure to disturbing content causes \textit{secondary traumatization} \cite{mccann1990vicarious} for moderators, often leading to post-traumatic stress disorder (PTSD) \cite{steiger-moderation}. This annotation is similar to volunteer moderation of identity-specific online communities, such as those studied by \citet{dosono-aapi-moderation}, who found that moderation constitutes significant emotional labor and often leads to burnout.  

%\subsection{LLMs for Data Annotation}

%\subsection{Model-created benchmarks}

\section{Methods}

\subsection{Jewish Community Survey}
Because one of the purposes of the WinoSemitism dataset is to validate the generalizability of the benchmark development approach from \citet{felkner-etal-2023-winoqueer}, we closely follow their methodology on survey design and human annotation of survey data. 
The Jewish community survey was deemed exempt by our institution's IRB.
Participants were recruited through a variety of channels, including researchers' personal networks, Jewish social media channels, synagogues, and Jewish student organizations. 
Interested respondents were directed to a screening questionnaire. Those who met the screening criteria (eighteen or older, English-speaking, and identifying as Jewish) then had the opportunity to review the informed constent document. 
The consent form explained the potential benefits and risks of the survey and informed participants that they could skip questions or leave the survey for any reason at any time during their participation. 

Participants then answered demographic questions. They were asked to self-identify their gender and ethnicity and provide their age range and country of residence. 
They were also asked about their Jewish cultural background (e.g. Ashkenazi, Sephardic, Mizrahi) and religious identity (e.g. Conservative, Reform). 
The demographic questions about Jewish identity were developed in consultation with an expert in Jewish Studies. 
These questions were included to ensure that Jewish people of diverse backgrounds and experiences were included in the WinoSemitism dataset. 

Following the demographic questions, participants answered multiple choice and free-response questions. 
The multiple choice questions listed a variety of common stereotypes about Jewish people. 
Participants were asked whether they thought each stereotype was positive, neutral, or negative, and whether they had personally experienced the stereotype. 
There were also open-ended questions for participants to write in additional stereotypes, both about Jews in general and about specific subcommunities with which they identified. 

The survey was active for about 10 weeks in late 2023. We had a total of 203 respondents. The overwhelming majority of respondents were white and from the United States. Respondents were relatively evenly distributed across age ranges. There were significantly more responses from women than men; we also saw a small number of responses from trans and nonbinary Jews. Respondents were mostly Ashkenazi, with very few Sephardi, Mizrahi, and Persian Jews represented. Respondents were relatively evenly distributed across religious identities, with Conservative and Reform Jews having the highest number of respondents. 

The vast majority of survey responses concerned general antisemitism, rather than stereotypes about specific subgroups of the Jewish community. 
There are not enough predicates about identity subgroups to report meaningful results; thus, we only report quantitative results aggregated over the entire dataset. 
This is significantly different than the results reported on the WinoQueer dataset, for which survey responses were much more evenly distributed. 

\subsection{WinoSemitism Benchmark Construction}
The WinoSemitism benchmark consists of pairs of \textit{stereotypical} (i.e. antisemitic) and \textit{counter-stereotypical} sentences. 
Each sentence is constructed from the following components: 
\begin{itemize}
    \item Sentence templates, which are the structures into which other components are placed. Sentence templates were constructed based on findings from \citet{cao-etal-2022-theory}.
    
    \item Identity descriptors, which are usually either ``Jewish people'' or ``Jews'' but sometimes include specific subpopulations, such as ``Jewish women'' or ``Orthodox Jews.''
    
    \item Counterfactual ID descriptors, which were \textit{Christian, secular, atheist, and nonreligious}.
    
    \item Common Jewish names, which were selected by an expert in Jewish studies (based on works including \citet{fermaglich-jewish-names}) from  the US Social Security Administration's list of common names \footnote{\url{https://www.ssa.gov/oact/babynames/decades/names2010s.html}} from 1970 to 2019. 
    
    \item Attested harm predicates, which were extracted from survey responses via annotation by expert humans. Most predicates apply to Jewish people in general, but some were specific to subgroups of the Jewish community. After extraction, predicates were manually edited to ensure syntactic correctness of constructed sentences.
    
\end{itemize}

Benchmark sentences were constructed by sampling from each component list.
Each sentence template was filled in with each extracted predicate and the predicate's corresponding identity descriptor(s). 
Then, the subjects of singular templates were filled in with he/him and she/her pronouns as well as a random sample of 5 masculine and 5 feminine names. 
This yielded the set of stereotypical sentences. 
Finally, for each stereotypical sentence, two of the four counterfactual identity descriptors were sampled and two sentence pairs were constructed.
Our use of random sampling of names and counterfactual identity descriptors is a departure from previous work, which constructed benchmarks using a strict Cartesian product of all component categories. 
However, we had a much larger list of names than previous work, and we chose to use random sampling to keep the overall size of the WinoSemitism benchmark roughly comparable to previous work. 
In total, the WinoSemitism benchmark consists of 58,816 sentence pairs. 
An example WinoSemitism sentence pair is \textit{``All Jews are greedy.''} and \textit{``All Christians are greedy.''}

\subsection{GPT Extraction}
The vast majority of human effort in the benchmark construction process described above is in the extraction of attested harm predicates from free-response survey data. 
In order to alleviate the financial and psychological burden of human annotation, we tested whether GPT-3.5-Turbo can be used to perform this predicate extraction. 
In our human annotation setup, annotators were presented with the full text of one response to one survey question, which could range in length from a short phrase to several sentences. 
Annotators were asked to extract any number of attested harm predicates and record each separately. 
Thus each response to each question is paired with a list of human-extracted predicates. 
Survey responses for which humans were unable to extract harm predicates were removed from the dataset. 

We performed GPT-3.5-Turbo predicate extraction for both the WinoQueer and WinoSemitism datasets, using the same prompting setup for both. 
For each survey answer, the model was prompted $N$ times, where $N$ is the number of ground truth predicates extracted by a human. 
This $N$ would be unknown in a realistic use case, so our experiments represent an artificially easy annotation task.
Temperature was set to 0.3 for all experiments. Full details of the prompts used can be found in Section~\ref{sec:appendix}.

\subsection{GPT Benchmark Construction}
After attested harm predicates have been automatically extracted from both LGBTQ+ and Jewish community survey responses using GPT-3.5-Turbo, they are then used to create model-assisted versions of the WinoQueer and WinoSemitism benchmarks. 
These datasets are named GPT-WinoQueer (GPT-WQ) and GPT-WinoSemitism (GPT-WS). 
To differentiate between versions, we refer to the original, human-created datasets as Human-WinoQueer (H-WQ) and Human-WinoSemitism (H-WS), respectively. 
To build the GPT-WQ and GPT-WS benchmarks, we use exactly the same lists of template sentences, identity descriptors, names, pronouns, and counterfactuals as in the corresponding human-extracted dataset. These component lists are combined with GPT-extracted predicates using exactly the same methods as orginal datasets (strict Cartesian product for WQ, Cartesian product with some random sampling for WS). This generation yielded 45,468 sentence pairs in GPT-WQ and 68,472 sentence pairs in GPT-WS. The GPT-extracted datasets differ slightly in size from the corresponding human-extracted datasets due to differing numbers of unique predicates extracted. 

The only difference between the H-W* and GPT-W* datasets is thus the quality of the extracted predicates. 
By comparing bias scores of the same models on the two pairs of datasets, we can determine whether the GPT-extracted bias scores correlate well with the human-extracted baselines. 
Thus, we can assess whether GPT performs comparably to humans on the highly context-sensitive task of bias benchmark construction.

\subsection{Evaluation Metrics}

\subsubsection{WinoSemitism Baseline}
All of our bias benchmark datasets (H-WS, GPT-WQ, and GPT-WS) use the same bias score as the original WinoQueer benchmark. Intuitively, the bias score is the percentage of sentence pair for which the tested model has a higher probability of predicting the stereotypical sentence  than the counterstereotypical sentence. An ideal bias score is 50, meaning that a model is equally likely to apply a stereotype to either the group targeted by the stereotype or the corresponding majority group. Model probabilities are calculated using the pseudo-log-likelihood score from \citet{nangia-etal-2020-crows}, which was extended to autoregressive models by \citet{felkner-etal-2023-winoqueer}. This bias score is used for both human-created and model-created benchmark datasets. The total computation budget for all LLM evaluations was around 500 GPU hours across NVIDIA P100s, V100s, and A100s.

\subsubsection{Auxiliary Evaluation of GPT-Extracted Predicates}
In addition to comparing WQ and WS bias scores, we use several auxilary metrics to evaluate the quality of GPT-extracted harm predicates. We use the following quantitative metrics:
\begin{itemize}
    \item \textbf{Exact match percentage:} percentage of GPT-extracted predicates that exactly match a human-extracted predicate for the same survey response. We expect this number to be very low, indicating that predicate extraction is a nontrivial natural language understanding task.
    \item \textbf{Cosine similarity score:} We take the cosine similarity of SBERT sentence embeddings from the \texttt{all-mpnet-base-v2} model \cite{reimers-2019-sentence-bert} for the human-extracted and GPT-extracted predicates. We consider embeddings of just the predicates (i.e. phrases) and of simple sentences containing the predicates. Higher cosine similarity scores indicate more similarity between human-extracted and GPT-extracted predicates. However, this metric is prone to overestimating similarity in cases where the two predicates share many words but differ significantly in meaning. 
\end{itemize}
For both metrics, we compare each human-extracted predicate to all GPT-extracted predicates for the same sentence and take the best score. 

In addition to quantitative similarity scores, we perform a qualitative human analysis to better understand the failure types. All GPT extracted predictates were reviewed \textit{post hoc} by researchers. Each was classified into one of five outcome categories: \textbf{Correct}, \textbf{Semantically Correct} (but syntactically invalid and in need of post-editing), \textbf{Opposite} (counter to the stereotype being attested), \textbf{Hallucination} (not present in the survey), and \textbf{Other} (incorrect but neither opposite nor hallucination).

\section{Results}

\subsection{WinoSemitism Baseline Results}
\begin{table}[ht]
\centering
\begin{tabular}{|l|c|}
\hline
\textbf{Model} & \textbf{WinoSem. Score}\\
\hline
BERT & 69.53 \\
 \hline
 RoBERTa & 66.51 \\
 \hline
 ALBERT & 65.27 \\
 \hline
 BART & 63.50 \\
 \hline
 gpt2  & 70.11 \\
 \hline
 BLOOM & 70.31 \\
 \hline
 OPT & 75.17 \\
 \hline
 \textbf{Mean, all models} & 69.03 \\
 \hline
\end{tabular}
\caption{WinoSemitism baseline results for 7 families, comprising 20 off-the-shelf language models; complete results are in Table~\ref{tab:ws_complete_results}. Scores over 50 indicate presence of antisemitic stereotypes in the model. All tested models show some degree of antisemitism.}
\label{tab:ws_results}
\end{table}

First, we present baseline results on the WinoSemitism dataset across 20 publicly available LMs. 
Following \citet{felkner-etal-2023-winoqueer}, we evaluate on BERT \cite{devlin-etal-2019-bert}, RoBERTa \cite{roberta}, ALBERT \cite{Lan2020ALBERT}, BART \cite{lewis-etal-2020-bart}, GPT-2 \cite{Radford2019LanguageMA}, BLOOM \cite{BLOOM}, and OPT \cite{zhang2022opt}
Results are summarized in Table \ref{tab:ws_results}. 
In general, the baseline results show that all 20 of the tested models show significant antisemitic bias, as defined by members of the Jewish community. 
On average, models applied community-defined antisemitic stereotypes to Jews in 69.03\% of cases and applied the same stereotypes in non-Jews in 30.97\% of cases. 
This means that models are more than twice as likely to apply antisemitic statements to Jews as they are to non-Jews. 

We also tested stereotypes about specific subgroups of the Jewish community. 
We received very few survey responses about subgroup-specific biases, so our test sets for most subgroups are too small to report meaningful quantitative results.
Anecdotally, however, we notice a few worrying trends. 
First, models stereotype Jewish women and mothers much more frequently than Jews in general. 
The average WinoSemitism score for Jewish mothers compared to non-Jewish mothers across all models is 76.7, and the average WinoSemitism score for Jewish women is 84.34. 
This suggests that, in addition to sexism and antisemitism measured in isolation, models are reproducing intersectional biases specific to Jewish women. 
Second, models reproduce stereotypes about Jews even more strongly in sentences related to Israel, Palestine, Zionism, and anti-Zionism, with bias scores in the mid-to-high nineties. Training data for LLMs most likely reflect the high degree of controversy present in global public discourse about Israel and Palestine, debates about whether anti-Zionism constitutes antisemitism, and conflation of the State of Israel with all Jews everywhere.

\subsection{Predicate Extraction}
\begin{table}[ht]
\centering
\begin{tabular}{|l|c|c|c|}
\hline
\textbf{ID Group} & \textbf{\% Exact} & \textbf{PCS} & \textbf{SCS}
\\
\hline
LGBTQ & 8.33 & \textbf{0.54} & \textbf{0.84} \\
Queer & 7.14 & 0.47 & 0.79 \\
Transgender & 3.23 & 0.44 & 0.75 \\
Nonbinary & \textbf{11.11} & 0.51 & 0.82 \\
Bisexual & 4.76 & 0.43 & 0.76 \\
Pansexual & \textit{0.00} & \textit{0.39} & \textit{0.71} \\
Lesbian & 5.88 & 0.42 & 0.74 \\
Asexual & \textit{0.00} & 0.40 & 0.74 \\
Gay & 5.36 & 0.52 & 0.83 \\
\hline
\textbf{WQ Overall} & 5.40 & 0.47 & 0.78 \\
\hline
\hline
\textbf{WS Overall} & 18.14 & 0.61 & 0.82 \\
\hline
\end{tabular}
\caption{GPT predicate extraction automated metric scores for LGBTQ+ identity subgroups, GPT-WQ overall, and GPT-WS overall. \% Exact is the percentage of cases where human- and GPT-extracted predicates match exactly. PCS is phrase cosine similarity of SBERT embeddings for just the extracted phrases. SCS is sentence cosine similarity of SBERT embeddings for dummy sentences containing  extracted phrases. For ID subgroups, \textbf{bold} represents best scores and \textit{italics} represents worst scores.}
\label{tab:cosine_sim_results}
\end{table}

Below, we present results for our experiments on automated extraction of attested harms from survey data.
We evaluate the success of our approach on both quantitative similarity metrics and qualitative human analysis. 
Table \ref{tab:cosine_sim_results} summarizes the quantitive similarity metrics for the GPT-WQ and GPT-WS datasets. 
We report GPT-WQ results by identity group, but GPT-WS results are only reported in aggregate due to very small test set sizes for most identity subgroups. 
As expected, we have very low rates of exact matches (disregarding capitalization and punctuation) between human-extracted and GPT-extracted predicates. 
This shows that the task is non-trivial and is more than simply selecting a span from the input text. 
We also see relatively low cosine similarity scores between the SBERT sentence embeddings of human-extracted and GPT-extracted predicates. 
These low cosine similarities mean that GPT is unable to adequately identify attested harms in survey responses.
While the cosine similarities and exact match rates are better for WinoSemitism than for WinoQueer predicates, the scores still indicated significant quality issues with model-extracted predicates. 
As in WQ baseline results, there is a disparate impact on subgroups marginalized within the LGBTQ+ community. 
Model-extracted predicates are especially bad for lesbian, bisexual, pansexual, and asexual individuals. 
These results suggest that using LLMs in the construction of bias benchmarks is likely to produce benchmarks that are generally low-quality and fail to accurately measure biases against multiply marginalized populations.
This runs the risk of exacerbating existing inequalities if too much trust is placed in benchmarks that underestimate bias effects on certain subgroups of marginalized communities. 

\begin{table}[ht]
\centering
\begin{tabular}{|l|c|c|c|c|}
\hline
\textbf{ID Group} & \textbf{\%C} & \textbf{\%S} & \textbf{\%O} & \textbf{\%H} \\
\hline
LGBTQ &  52.78 & 2.78 & 8.33 & 30.56 \\
Queer & 35.71 & 25.00 & 7.14 & 32.14 \\
Transgender & 35.48 & 22.58 & 6.45 & 32.36  \\
Nonbinary & 33.33 & 14.81 & 3.70 & 48.15 \\
Bisexual & 31.00 & 21.43 & 2.38 & 42.86 \\
Pansexual & 17.65 & 35.29 & 0.00 & 41.18 \\
Lesbian & 35.29 & 17.65 & 29.41 & 11.76 \\
Asexual & 12.50 & 50.00 & 8.33 & 25.00 \\
Gay & 57.15 & 19.64 & 19.64 & 1.79 \\
\hline
\textbf{WQ Overall} & 38.13 & 21.58 & 6.12 & 31.30 \\
\hline
\end{tabular}
\caption{Human evaluation results for subgroups of GPT-WinoQueer dataset. \%C is the percentage of correct extractions. \%S is the percentage of semantically correct model responses requiring only grammatical corrections. \%O is the percentage of model responses expressing the opposite of the input text. \%H is the percentage of cases where the model hallucinates a response that is not in the input. Respones categorized as ``other'' are omitted for brevity; $< 6\%$ 
 in all cases.}
 \label{tab:gptwq_human}
\end{table}

\begin{figure}[ht]
    \centering
    \includegraphics[width=0.5\textwidth]{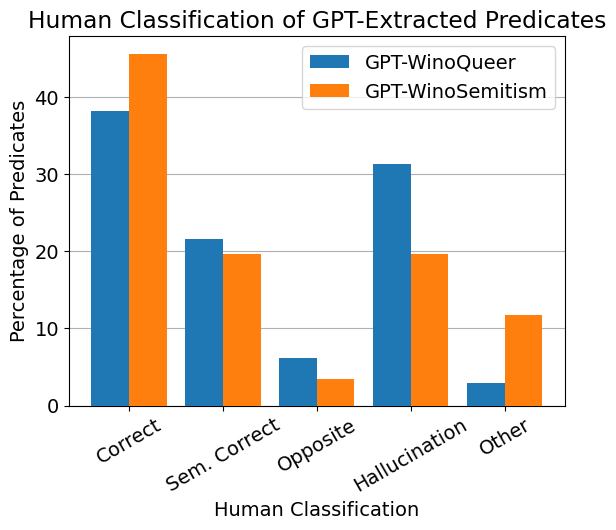}
    \caption{Comparison of human post-evaluation results for GPT-WinoQueer (blue, left) and GPT-WinoSemitism (orange, right) datasets. GPT-WQ evaluation results are generally worse, with a lower proportion of correct responses and higher proportions of grammatically incorrect, opposite, and hallucinated responses.}
    \label{fig:human_class_comp}
\end{figure}

We also perform a \textit{post hoc} human analysis of GPT-extracted predicates. Results from this analysis are summarized in Fig.~\ref{fig:human_class_comp}. 
We observe that, for both GPT-WQ and GPT-WS, less than half of GPT-extracted predicates are classified correct, i.e. immediately usable for benchmark construction. 
Around twenty percent of model-extracted predicates require only syntactic, not semantic, changes. 
While it is possible that some of this editing could be automated, it is likely that much of it would require human correction or supervision to create a high-quality benchmark.

The first issue with model-extracted predicates is incomplete extractions. 
These are included in the \textit{other} category of Fig.~\ref{fig:human_class_comp}. 
While a few such outputs are to be expected from both human and model annotation, the prevalence of incomplete extractions in the GPT-WinoSemitism dataset is concerning. 
In particular, many extractions consist of single verbs and are missing the nouns that are an important component of the attested stereotype. 
For example, the stereotypes ``Jews control the US government'' and ``Jews consider themselves superior to others'' are collapsed to ``Jews control'' and ``Jews believe,'' respectively. 
These predicates are unspecific and do not adequately reflect the harms attested by survey respondents. 

Another significant issue with model-extracted predicates is significant number of cases where the model extracts the \textbf{opposite} of the stereotype expressed in the survey response. 
This occurs in 6.12\% of GPT-WQ extractions and 3.43\% of GPT-WS extractions. 
While these are relatively small percentages, the downstream impact is that a significant portion of sentence pairs in the GPT-W* datasets are not only ignorning but directly contradicting the experiences of affected community members. 
Additionally, because of their surface-level similarity to correct extractions, these negation errors are not easily flagged by automated metrics like cosine similarity. 
On the WinoQueer survey response, we observe that many cases of negation concern instances of violence and harassment experienced by community members. 
Instead of correctly identifying the survey respondent as a victim, model-extracted predicates conflate the survey respondent with the \textit{perpetrator}. 
This impact is most severe for lesbian respondents (nearly 30\% of all model outputs), many of whom reported being catcalled and sexually objectified. 
GPT extraction incorrectly yielded benchmark sentences like ``Lesbians catcall others'' and ``All lesbians objectify others.'' 
While both of these are expressions of the harmful stereotype that lesbians are predatory towards straight women, neither of them was actually present in the survey data, so they do not meet the criteria for inclusion in a \textit{community-sourced} fairness benchmark.
Benchmarks that spuriously accuse marginalized people of perpetrating the violence of which they report being victims is a serious risk of automated extraction of attested harms for benchmark construction.

Most concerning, however, is the extremely high rate of model hallucinations in extracted predicates. 
Over thirty percent of GPT-WQ predicates, and nearly twenty percent of GPT-WS predicates, express a stereotype or bias that is not present in the survey response from which they were ``extracted.'' 
Most of these hallucinations are highly repetitive, suggesting that the model defaults to a very small set of priors when unable to extract a predicate from the input data. 
Repetitive hallucinations include ``are confused,'' ``struggle with mental health,'', and ``are promiscuous''  for LGBTQ+ survey responses. 
For Jewish survey responses, repeated hallucinations include ``are greedy'' and ``are manipulative.'' 
In both cases, even non-repeated hallucinations are still similar in meaning to repetitive hallucinations. 
These repetitive phrases are also in some model outputs that are classified as correct. 
First, this implies that in some of these cases, the model is not correctly performing the task on the input, instead relying on a limited set of priors; this indicates that the ``correct'' percentages might overestimate model capability.
Second, this has the effect of collapsing many attested harms into very few predicates. 
For example, in the GPT-WQ data, the model extracted either ``are promiscuous'' or ``engage in promiscuous behavior'' for the majority of survey responses mentioning any stereotypes about queer sexuality. 
This conflation of survey responses oversimplifies and ignores valuable community input. 

These repetitive hallucinations do express real and offensive stereotypes, which undoubtedly should be included in bias benchmarks. However, the main advantage of community-sourced bias benchmarks is the fact that they are grounded in the lived experience of affected community members. Using models to ``extract'' attested harms that are not actually present in survey responses negates the purpose of community-engaged bias definitions, effectively undoing the work of both researchers and community members in eliciting and sharing experienced harms.  

\subsection{Comparing Human- and GPT-Extracted Benchmarks}

\begin{figure}[ht]
    \centering
    \includegraphics[width=0.5\textwidth]{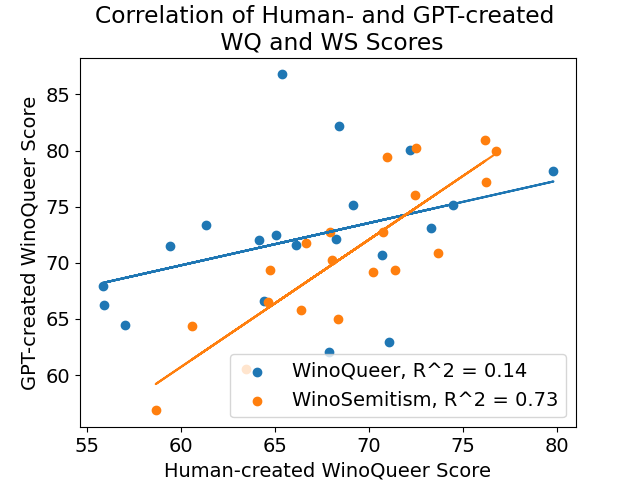}
    \caption{GPT-W* vs H-W* bias scores and best fit lines. These bias scores should be strongly linearly correlated. However, we observe extremely weak correlation for WQ and only moderate correlation for WS. In both cases, the GPT-created dataset is a very poor proxy for the human-created dataset.}
    \label{fig:wq-and-ws-correl}
\end{figure}

Ideally, human-created and model-created benchmarks should be measuring the same underlying model behaviors and should both score a given model similarly. 
Equivalently, there should be a strong, positive linear correlation between per-model H-W* and GPT-W* scores. 
Fig.~\ref{fig:wq-and-ws-correl} shows the poor correlation between scores on the two benchmarks. 
For H-WinoQueer and GPT-WinoQueer, the $R^2$ value is only 0.14, indicating a very weak correlation. 
The WinoSemitism datasets have a moderate correlation ($R^2 = 0.73$), but we would still expect a stronger correlation if the two benchmarks were well-calibrated with respect to each other. 
GPT-created bias benchmarks are a very poor approximation for the corresponding human-created benchmarks. 

Just as the quality of model extractions is inequitably distributed across subgroups of the LGBTQ+ community, poor calibration of GPT-created bias benchmarks also has disparate impacts on marginalized subcommunities. 
Of the nine LGBTQ+ identity subgroups considered, six had $R^2 < 0.5$, including $R^2 = 0$ for lesbians. 
Model-created bias benchmarks perform even more poorly for marginalized subcommunities, falsely underestimating the likelihood of intersectional stereotypes in model outputs.

\section{Conclusions}

This paper introduced the WinoSemitism dataset, a \textit{community-sourced} benchmark dataset for measuring bias against Jewish people in large language models (LLMs). 
We also investigated the feasibility of using a large language model to assist humans with the difficult, emotionally taxing work of developing LLM fairness benchmarks.  
In particular, we used GPT-3.5-Turbo to extract \textit{attested harm predicates} from community survey responses. 
Even on this limited subtask with additional information that would be unavailable to a model in a realistic scenario, we find that GPT-3.5-Turbo has unacceptably poor performance on attested harm extraction. 
We present quantitative and qualitative evidence of serious quality issues with model-extracted predicates, including high rates of both misrepresenting survey responses and hallucinating harms not present in input text. 
These quality issues replicate many of the problems of \textit{crowd-sourced} (i.e. not community-sourced) fairness benchmarks, including lack of specificity, nonsensical stereotypes, and ungrammatical sentences, all of which would require considerable human effort to correct.

We also show that model-created benchmarks are poorly correlated with human-created benchmarks from the same underlying community surveys, indicating that GPT-created benchmarks are not measuring the same model behavior as human-created benchmarks and are poorly grounded in the actual experiences of marginalized communities. 
Finally, the impacts of poor extraction quality disparately impacts already-marginalized subgroups of affected communities, meaning that model-created benchmarks are likely to replicate and exacerbate existing intracommunity inequality.
Thus, we conclude that using LLMs to process human survey responses into fairness benchmark datasets yields unnacceptably low-quality benchmarks and largely negates the positive impacts of conducting a community survey in the first place. 
Annotations from expert humans with lived experience as members of the relevant community are absolutely essential to the construction of high-quality LLM fairness benchmarks. 
Given these findings, we strongly caution against the use of LLMs in the creation of fairness benchmarks intended to evaluate LLMs and stress the important of human annotators, especially in sensitive and highly context-dependent tasks. 

\section{Limitations}

\subsection{Jewish Community Survey and WinoSemitism Dataset}
Our survey of Jewish community members, and by extension the WinoSemitism dataset, have several important limitations. 
First, our sample is relatively small, and is not likely to be representative of the global Jewish community.
The WinoSemitism dataset is also not representative of all forms of antisemitism or all stereotypes about Jewish people. 
There are undoubtedly antisemitic stereotypes we did not include in the study because they were not reported by our small sample of participants. 
For practical reasons, sample participants were limited to English-speaking adults 18 and older, so the views of Jewish youth and Jews who do not speak English are not represented in this sample.

\subsection{GPT-WQ and GPT-WS Dataset Construction}
Our construction of the GPT-WQ and GPT-WS datasets also has important limitations. For predicate extraction, our prompts were developed empirically, and we have not rigorously tested all possible prompts. 
Therefore, it is possible that a different prompt, or even the same prompt given to a different release of GPT-3.5-Turbo, would produce different results than those reported in this paper. 
We also did not test all current LLMs, so it is possible that other models will perform better than GPT-3.5-Turbo on this annotation tasks. 
However, we believe our conclusions about the serious quality issues caused by LLM use in benchmark construction are not solely limited to a specific prompt and model release. 

In constructing the GPT-WQ and GPT-WS benchmark sets, we closely follow the methodology of \citet{felkner-etal-2023-winoqueer}. 
Their methodology has several limitations, which are discussed in detail in their work. 
Important limitations affecting this work include low syntactic diversity of benchmark sentences and inability to rigorously examine intersectional biases. 
While the H-WQ and GPT-WQ datasets do not adhere to a strict understanding of gender as binary, current versions of the H-WS and GPT-WS do treat gender as binary. 
They use only she/her and he/him pronouns, which are mapped to traditionally feminine and traditionally masculine names, respectively. 
This is an area for improvement in future versions of the dataset.

\subsection{Evaluation}
The first important limitation of our evaluation methods is our choice of models on which to evaluate. 
Because our survey was conducted only in English, we evaluate the resulting datasets on monolingual English language models. 
The 20 models on which we evaluated are only a very small fraction of the language models available today. 
They are also small in terms of number of parameters.
We evaluate on models up to seven billion parameters, but many state-of-the-art models have hundreds of billions of parameters.
Our evaluation metric is limited to open-source models for which we can directly access token probabilities. 
It has not yet been expanded to closed-source models that are accessible only via prompting. 

% no acks in submission
% \section*{Acknowledgements}

\bibliography{anthology,custom}

\appendix

\section{Prompt Details}
\label{sec:appendix}

\lstset{%
  language=[LaTeX]TeX,
  backgroundcolor=\color{gray!25},
  basicstyle=\ttfamily,
  breaklines=true,
  columns=fullflexible,
  caption={Prompt for GPT Predicate Extraction}, 
  alsoletter={<},
}

For all experiments, the basic template prompt was:
\begin{lstlisting}
From the following sentence, extract a statement of one discrete stereotype about <IDENTITY NOUN>. 
Stereotypes should be in the form of a predicate that would fit in the sentence 'All <IDENTITY NOUN> <predicate>.' 
The predicate should start with a verb and SHOULD NOT inlcude individual characters separated by commas.
Do not respond with a full sentence.
There may be multiple stereotypes in the sentence, so extract one that is not already in the list.

Sentence: <SENTENCE>

Previously Extracted: <PREV>

Predicate: 
\end{lstlisting}

\texttt{<SENTENCE>} was replaced with a a single survey response. \texttt{<IDENTITY NOUN>} was replaced with an identity descriptor matching the survey respondent. \texttt{<PREV>} was replaced with an empty string at first, and then each model was prompted iteratively $N$ times, where $N$ is the number of ground truth predicates extracted by human annotators. Each iteration with the same input sentence, the model's previous response was added to the list of previously extracted responses. 

\section{Complete WinoSemitism Baseline Results}

\begin{table}[ht]
\centering
\begin{tabular}{|l|c|}
\hline
\textbf{Model} & \textbf{WinoSem. Score}\\
\hline
BERT-base-unc &  70.20 \\
 BERT-base-cased &  72.44 \\
 BERT-lg-unc & 64.74 \\
 BERT-lg-cased & 70.74 \\
 \hline
 RoBERTa-base & 64.65 \\
 RoBERTa-large & 68.37 \\
 \hline
 ALBERT-base-v2 & 58.66 \\
 ALBERT-large-v2 & 63.47 \\
 ALBERT-xxl-v2 & 73.67 \\
 \hline
 BART-base & 66.4 \\
 BART-large & 60.59 \\
 \hline
 gpt2 & 67.94 \\
 gpt2-medium & 70.99 \\
 gpt2-xl & 71.4 \\
 \hline
 BLOOM-560m & 66.67 \\
 BLOOM-3b & 76.21 \\
 BLOOM-7.1b & 68.04 \\
 \hline
 OPT-350m & 72.49 \\
 OPT-2.7b & 76.26 \\
 OPT-6.7b & 76.75 \\
 \hline
 \textbf{Mean, all models} & 69.03 \\
 \hline
\end{tabular}
\caption{WinoSemitism baseline results for 20 off-the-shelf language models. Scores over 50 indicate presence of antisemitic stereotypes in the model. All tested models show some degree of antisemitism.}
\label{tab:ws_complete_results}
\end{table}

Complete results for all 20 tested models on the human-annotated WinoSemitism dataset are listed in Table~\ref{tab:ws_complete_results}.
\end{document}